\documentclass[11pt,a4paper]{article}
\usepackage[utf8]{inputenc}
\usepackage[dvipdfmx]{graphicx}
\usepackage{amssymb}
\usepackage{times}
\usepackage{latexsym}
\usepackage{arydshln}
\usepackage{color}
\usepackage{stmaryrd}
\usepackage{multirow}
\usepackage{comment}
\usepackage[hyperref]{acl2021}
\usepackage{gb4ea}

\newcommand{\setof}[1]{\ensuremath{\left \{ #1 \right \}}}
\newcommand{\SR}[1]{\ensuremath{\mathsf{#1}}}
\newcommand{\Subj}{\SR{subject}}
\newcommand{\Pred}{\SR{predicate}}
\newcommand{\Obj}{\SR{object}}
\newcommand{\Action}{$\langle$\Subj, \Pred, \Obj$\rangle$}
\newcommand{\ActionLabel}[3]{$\langle$\SR{#1}, \SR{#2}, \SR{#3}$\rangle$}
\newcommand{\ActionLabeld}[3]{$\langle$#1, \SR{#2}, \SR{#3}$\rangle$}
\newcommand{\ActionLabelt}[2]{$\langle$#1, \SR{#2}, $\phi\rangle$}

\newcommand{\todo}[1]{\textcolor{black}{#1}}

\usepackage{microtype}

\aclfinalcopy
\def\aclpaperid{11}

\title{Building a Video-and-Language Dataset with Human Actions \\ for Multimodal Logical Inference}

\author{Riko Suzuki$^1$\\
  {\small \tt suzuki.riko@is.ocha.ac.jp}\And
  Hitomi Yanaka$^{2}$\\
  {\small \tt hyanaka@is.s.u-tokyo.ac.jp}\AND
  Koji Mineshima$^3$\\
  {\small \tt minesima@abelard.flet.keio.ac.jp}\And
  Daisuke Bekki$^1$\\
  {\small \tt bekki@is.ocha.ac.jp}\AND
  $^1$\mbox{\rm Ochanomizu University, Tokyo, Japan}\\
  $^2$\mbox{\rm The University of Tokyo, Tokyo, Japan}\\
  $^3$\mbox{\rm Keio University, Tokyo, Japan}
}

\date{}

\begin{document}

\maketitle

\begin{abstract}
This paper introduces a new video-and-language dataset with human actions for multimodal logical inference, which focuses on intentional and aspectual expressions that describe dynamic human actions.
The dataset consists of 200 videos, 5,554 action labels, and 1,942 action triplets of the form \Action\ that can be translated into logical semantic representations.
The dataset is expected to be useful for evaluating multimodal inference systems between videos and semantically complicated sentences including negation and quantification.
\end{abstract}

\section{Introduction}

Multimodal understanding tasks~\cite{clevr,nlvr,nlvr2} 
have attracted rapidly growing attention from both computer vision and natural language processing communities, and various multimodal tasks combining visual and linguistic reasoning, such as visual question answering~\cite{vqa,tallyqa} and image caption generation~\cite{image-caption-generation}, 
have been introduced.
With the development of the multimodal structured datasets such as Visual Genome~\cite{visual-genome}, recent studies have been tackling a complex multimodal inference task such as Visual Reasoning~\cite{nlvr2} and Visual-Textual Entailment (VTE)~\cite{suzuki2019,snli-ve},
a task to judge if a sentence is true or false under the situation described in an image.

The recently proposed multimodal logical inference system~\cite{suzuki2019} uses first-order logic (FOL) formulas as unified semantic representations for text and image information.
The FOL formulas are structured representations that capture not only objects and their semantic relationships in images but also those complex expressions including negation, quantification, and numerals.
When we consider extending the logical inference system between texts and images to that between texts and videos, it is necessary to handle the property of video information: there are dynamic expressions to capture human actions and movements of things in videos more than in images.

\begin{figure}[t]
    \includegraphics[width=\linewidth]{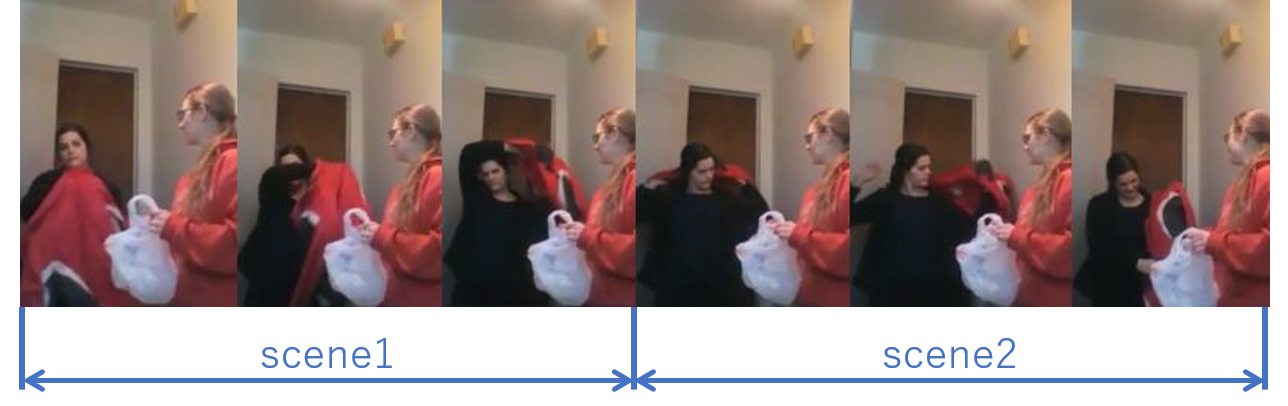}
    \caption{Inference example between a video and sentences. The description of this video is: \textit{The woman tried to put on her outerwear though she could not, because its zipper was not open completely.}}
    \label{fig:put_on}
\end{figure}

As an example, consider a video-and-language inference example in Figure \ref{fig:put_on}.
This video consists of \textsc{scene1},
where the sentence \textit{The woman puts on her outerwear} is true,
and \textsc{scene2}, where the sentence \textit{The woman takes off her outerwear} is true.
Note that the entire video represents richer information as expressed by the sentence \textit{the woman tries to put on her outerwear}.
To judge whether this sentence is true,
it is not enough to simply combine two actions,
\textit{putting on outerwear} and \textit{taking off outerwear}.
To capture this dynamic aspect of human action, it is necessary to take into account the information expressed by intentional phrases such as \textit{trying to put on outerwear}.

Towards such a complex multimodal inference between video and text, we build a new Japanese video-and-language dataset with human actions.
We annotate videos with action labels written in triplets of the form \Action,
where $\mathsf{object}$ can be empty (indicated by $\phi$).
Action labels contain not only basic expressions such as \ActionLabelt{\SR{person}}{run} and \ActionLabel{person}{hold}{cup}, but also expressions including intentional phrases such as \ActionLabel{person}{try\ to\ eat}{food}.
An advantage of using triplets \Action\ is that a triplet itself can serve as the semantic representation of a video and can be translated into logical formulas (see Section \ref{sec:SR}).
This paper introduces a method to create a video-and-language dataset involving aspectual and intentional phrases.
We collect a preliminary dataset labeled in Japanese for human actions.
We also analyze to what extent our dataset contains various aspectual and intentional phrases.
\todo{Our dataset will be publicly available at 
\url{https://github.com/rikos3/HumanActions}.}

\section{Related Work}
There have been several efforts to create human action video datasets in the field of computer vision.
Charades~\cite{charades} contains 9,848 videos of daily activities annotated with free-text descriptions and action labels in English.
Charades STA~\cite{charades-sta} is a dataset built by adding sentence descriptions with start and end times to the Charades dataset.
For Japanese video datasets, STAIR Actions~\cite{stair-action} is a dataset that consists of 63,000 videos with action labels.
Each video is about 5 seconds and has a single action label from 100 action categories.
Action Genome~\cite{action-genome} is a large-scale video dataset built upon the Charades dataset, which provides action labels and spatio-temporal scene graphs.

VIOLIN~\cite{violin} introduces a multimodal inference task between text and videos: given a video with aligned subtitles as a premise, paired with a natural language hypothesis based on the video content, a model needs to judge whether or not the hypothesis is entailed by the given video.
The VIOLIN dataset mainly focuses on conversation reasoning and commonsense reasoning, and the dataset contains videos collected from movies or TV shows.

Compared to the existing datasets, 
our dataset is distinctive in that action labels are written in structured representations \Action\ and contain various expressions such as \textit{continue to eat} and \textit{try to close} that support complex inference between videos and texts.

\section{Semantic Representations of Videos}
\label{sec:SR}

\citet{suzuki2019} proposed FOL formulas as semantic representations of text and images.
They use the formulas translated from FOL structures for images to solve a complex VTE task.
We extend this idea to semantic representations of videos.

FOL structures (also called first-order \textit{models}) are used to represent
semantic information in images~\cite{grim}.
An FOL structure for an image is a pair $(D, I)$
where $D$ is a domain consisting of
all the entities occurring in the image, and 
$I$ is an interpretation function that describes the attributes and relations holding of the entities in the image~\cite{suzuki2019}.

To extend FOL structures for images to those for videos, we add to FOL structures a set of scenes $S = \setof{s_1, s_2, \dots, s_n}$ that makes up a video, ordered by the temporal precedence relation.
This structure may be considered as
a possible world model for standard temporal logic~\cite{venema2001,Blackburn2002-BLAML-2}.
Thus, a video is represented by $(S, D, I)$, where $S$ is a set of scenes
linearly ordered by the temporal precedence relation,
$D$ is a domain of the entities, which is constant in all scenes, and $I$ is an interpretation function that assigns attributes and relations to the entities in each scene.
We assign personal IDs ($\mathsf{d_1}, \mathsf{d_2}, \dots, \mathsf{d_n}$) to people appearing in each scene.
Since the purpose of our dataset is to label human actions, we assign IDs to
people, but not to non-human objects.

To facilitate the annotation of the attributes and relations holding of the entities in each scene, we use triplets of the form \Action\ 
given to each scene $\mathsf{s_i}$
as action labels, where \Obj\ may be empty.
This form itself can be seen as a semantic representation of videos. Furthermore, it can also be translated into an FOL formula,
in a similar way to the standard translation of modal logic to FOL~\cite{Blackburn2002-BLAML-2}.
The following examples show a translation from triplets in scenes into FOL formulas.

\begin{exe}
    \ex \label{ex:run}
    $\mathsf{s_1}: $\ActionLabelt{$\mathsf{d_1}$}{run} \\
    $\Rightarrow \SR{run}(\mathsf{s_1},\mathsf{d_1})$
    \ex \label{ex:hold_pillow}
    $\mathsf{s_2}$: \ActionLabeld{$\mathsf{d_1}$}{hold}{pillow}\\
    $\Rightarrow \exists x (\SR{pillow}(\mathsf{s_2},x) \land \SR{hold}(\mathsf{s_2},\mathsf{d_1},x))$
\end{exe}

\noindent
Here each predicate has an additional argument for a scene variable.
(\ref{ex:run}) means that the entity $\mathsf{d_1}$ runs in scene $\mathsf{s_1}$; (\ref{ex:hold_pillow}) means that the entity $\mathsf{d_1}$ holds a pillow in scene $\mathsf{s_2}$.

Each triplet can be translated into an FOL formula by using this method and thus serve as a semantic representation of a video usable in the semantic parser and inference system for the VTE task presented in \citet{suzuki2019}.
Though it is left for future work,
the dataset in which each scene of a video is annotated with triplets will be useful to evaluate the VTE system for videos.

\section{Data Collection}

\subsection{Video Selection}
We selected videos from the test set of the Charades dataset~\cite{charades}.
The Charades dataset contains videos drawing daily activities in a room such as \textit{drinking from a cup}, \textit{putting on shoes}, and \textit{watching a laptop or something on a laptop}.
Each video is collected via crowdsourcing: workers are asked to generate the script that describes daily activities and then to record a video of that script being acted out.

\begin{figure}[!t]
    \centering
    \includegraphics[scale=0.27]{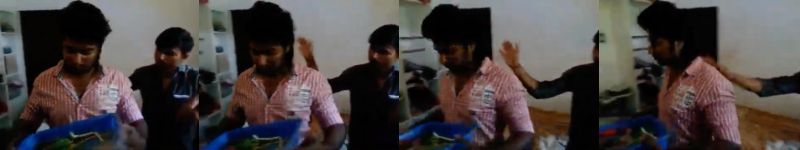}
    \caption{Example video for the action of \textit{touching someone's shoulder} from the Charades dataset.}
    \label{fig:tap}
\end{figure}

\todo{We select videos where multiple persons appear from the Charades test set to cover various actions within human interaction such as \textit{touching someone's shoulder} or \textit{handing something}.
These actions are expected to be described
in expressions involving various linguistic phenomena.
To collect videos where multiple persons appear,
we selected 200 videos whose descriptions include phrases \textit{another person}, \textit{another people}, and \textit{they}.}
Figure~\ref{fig:tap} shows a video example involving human interaction.

\subsection{Annotation}
We annotate each video with \Action\ triplet format as action labels that represent human-object activities.
\todo{We also annotate each action label with a start and end time to locate the activity accurately.
We ask two workers to freely write predicates and object names that describe human activities to collect various expressions.
Using this format the workers can freely decide the span of each scene and thus annotate a video with action labels more easily and flexibly.
In Section \ref{ssec:conv} below, we will explain how to convert the triplet action format with start and end times to FOL structures extended with scenes as presented in Section~\ref{sec:SR}.}

\paragraph{Subject}
We assign personal IDs ($\mathsf{d_1}, \mathsf{d_2}, \mathsf{d_3}, \dots$) to people in order of appearance in the video.
If multiple persons appear for the first time in the same scene, we assign personal IDs to people appearing in order from left to right.

\paragraph{Predicate}\
In a triplet,
\Pred\ contains various expressions such as aspectual and intentional phrases for describing dynamic human actions in videos, those phrases that do not usually appear in captions for static images.
The following examples show characteristic predicates of videos.

\vspace{-0.5em}
\begin{itemize}
    \setlength{\parskip}{0.1em}
    \setlength{\itemsep}{0.1em}
    \item predicates for utterance and communication
    \\(e.g. \textit{speak}, \textit{talk}, \textit{tell}, \textit{ask}, \textit{listen})
    \item predicates for intention and attitude \\(e.g. \textit{try to eat}, \textit{try to close}).
    \item aspectual predicates \\(e.g. \textit{start talking}, \textit{continue to eat})
\end{itemize}
\vspace{-0.5em}

\noindent
We allow workers to use not only a transitive or intransitive verb but also verb phrases for predicates such as \textit{try to V} and \textit{continue to V} to collect diverse aspectual and intentional phrases.

\paragraph{Object}

\begin{figure}[h]
    \centering
    \includegraphics[scale=0.27]{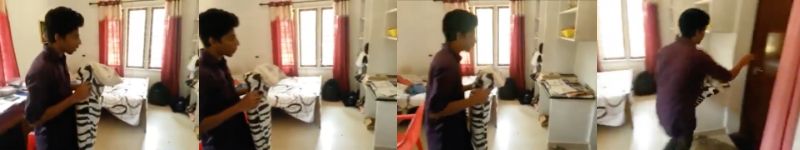}
    \caption{\textit{A man is running while holding a pillow.} Action labels are \ActionLabeld{$\mathsf{d_1}$}{hold}{pillow}\ and \ActionLabelt{$\mathsf{d_1}$}{run}}
    \label{fig:run}
\end{figure}

The \Obj\ in a triplet contains an object name or personal ID.
If the item in \Pred\ is an intransitive verb, \Obj\ is empty.
For instance,
in Figure~\ref{fig:run}, the object for the predicate \SR{hold} is \SR{pillow} and the object is empty for the predicate \SR{run}.

\subsection{Validation}
\todo{In this work, we ask three workers to either annotate or merge action labels.
All of the workers are native speakers of Japanese.
We merge and confirm action labels in the following steps:
(1) merge action labels made by two workers and arrange them in ascending order of start times, (2) watch videos by three workers to see if an action label is correct, and (3) if action labels duplicate, select one action label.}

\begin{table*}[!t]
    \centering
    \scalebox{0.8}{
    \begin{tabular}{lcccccc}
       \hline
       \textbf{Dataset} & \textbf{Videos} & \textbf{Average time} & \textbf{Average of} & \textbf{Action} & \textbf{English} & \textbf{Japanese}
       \\
        &  & \textbf{(sec)} & \textbf{action labels} &  \textbf{categories}& & 
        \\
       \hline
       Charades~\cite{charades} & 9848 & 30 & 6.8 & 157 & \checkmark &
       \\
       ActionGenome~\cite{action-genome} & 9848 & 30 & 170 & 157 & \checkmark &  
       \\
       STAIR Actions~\cite{stair-action} & 102462 & 5-6 & 1.0 & 100 & \checkmark &\checkmark 
       \\
       \hline
       \textbf{Ours} & 200 & 30 & 27.77 & \textbf{1942} & &\checkmark
       \\
       \hline
    \end{tabular}
    }
    \caption{A comparison of our dataset with existing datasets}
    \label{tab:compare}
\end{table*}

\begin{figure*}
    \centering
    \includegraphics{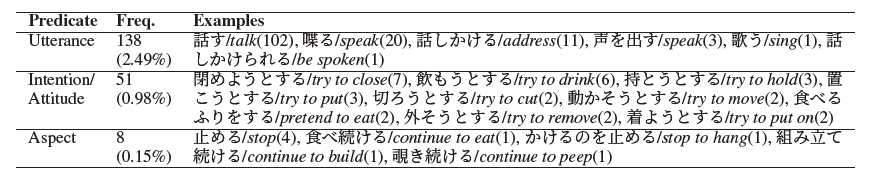}
    \caption{Predicates for utterance, intention and aspect}
    \label{tab:verbtypes}
\end{figure*}

\begin{figure*}[!ht]
\begin{center}
    \includegraphics[width=\linewidth]{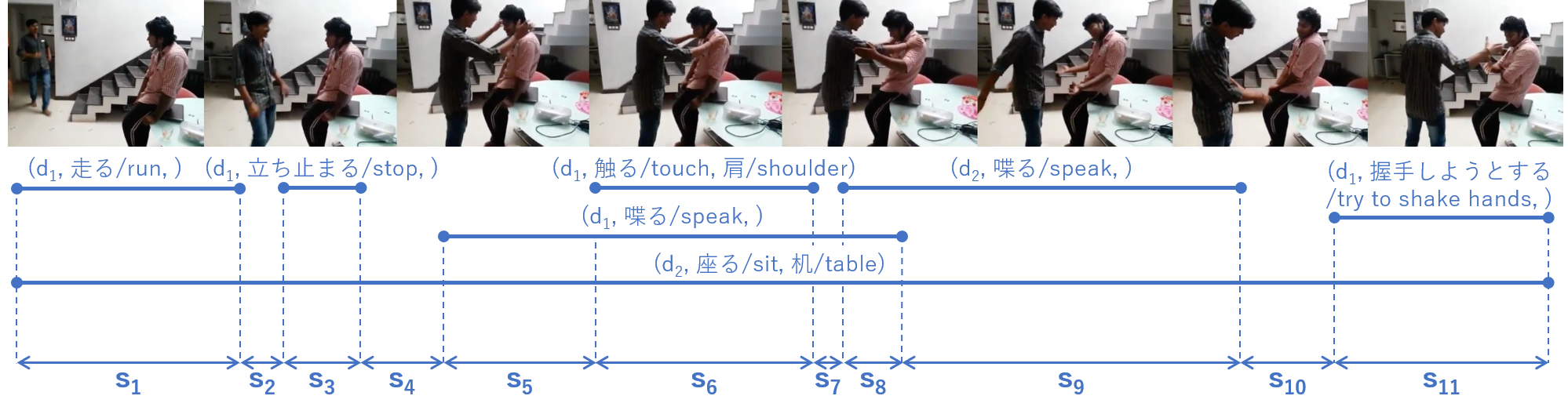}
\caption{Annotation example of a video labeled with various types of predicates. Here $s_1, \ldots, s_{11}$ are scenes linearly ordered by the temporal precedence relation.}
\label{fig:examples}
\end{center}
\end{figure*}

Regarding duplicated action labels, the labels and their start and end time are determined according to the agreement of three workers.
Consider the following duplicate case.

\begin{center}
\begin{tabular}{lll}
$(\mathsf{\sigma_1})$ & 0:10-0:13 & \ActionLabeld{$\mathsf{d_1}$}{hold}{clothes} \\
$(\mathsf{\sigma_2})$ & 0:11-0:14 & \ActionLabeld{$\mathsf{d_1}$}{hold}{clothes} \\
$(\mathsf{\sigma_3})$ & 0:11-0:15 &  \ActionLabeld{$\mathsf{d_1}$}{hold}{outerwear}
\end{tabular}
\end{center}

\noindent
In this case, $(\mathsf{\sigma_1})$ and $(\mathsf{\sigma_2})$ are duplicates in that \Subj, \Pred, and \Obj\ are the same while the start time and end time are different.
If the third worker judges that $(\mathsf{\sigma_2})$ is more adequate than $(\mathsf{\sigma_1})$, we
merge $(\sigma_1)$ and $(\sigma_2)$
and obtain the action labels below.

\begin{center}
\begin{tabular}{lll}
$(\mathsf{\sigma_1}')$ & 0:10-0:14 & \ActionLabeld{$\mathsf{d_1}$}{hold}{clothes} \\
$(\mathsf{\sigma_2}')$ & 0:11-0:15 &  \ActionLabeld{$\mathsf{d_1}$}{hold}{outerwear}
\end{tabular}
\end{center}

\subsection{Collection Statistics}

Table~\ref{tab:compare} shows that despite its size, our dataset contains more action categories than other previous datasets.
About 65\% of total action labels are action labels that appear only once.
This indicates that there are a wide variety of expressions.

\begin{figure}
    \centering
    \includegraphics{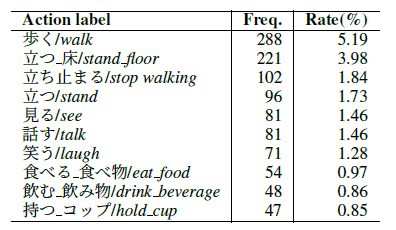}
    \caption{Top 10 frequent action labels. Action labels are written in form of \Pred\_\Obj\ or \Pred.}
    \label{tab:frequency}
\end{figure}

The dataset contains characteristic expressions of videos such as \textit{walk}, \textit{talk}, and \textit{stop walking}.
Table~\ref{tab:verbtypes} shows the frequency and examples of three types of predicates, i.e.,
utterance, intentional, and aspectual predicates.
The distribution of characteristic predicates of videos in our dataset was: 2.49\% predicates for utterance, 0.98\% predicates for intention and attitude, and 0.15\% aspectual predicates.
One possible reason for the low frequency of aspectual predicates is that Charades contains 30-second videos, which might be too short to describe multiple actions involving aspectual phrases.
It would be expected to increase the number of aspectual predicates if we annotate longer videos such as the VIOLIN dataset~\cite{violin},
which is left for future work.
The number of overlaps of action categories between ours and STAIR Actions~\cite{stair-action} is 28.
These results indicate that our dataset contains
more diverse action categories
compared to other datasets.

Table~\ref{tab:frequency} shows frequent action labels in our dataset.
Our dataset contains not only predicates for utterance, intention, and aspect, but also punctual verbs (e.g. \textit{stop walking} and \textit{turn on}) and durative verbs (e.g. \textit{sit} and \textit{wait}).

\subsection{Conversion to FOL structures}
\label{ssec:conv}

The triplet action forms with start and end points used in the annotation can be converted to 
FOL structures extended with scenes
presented in Section \ref{sec:SR}.
In the extended FOL structures, each scene is linearly ordered by the temporal precedence relation
and is uniquely characterized by the set of all the attributes and relations holding in it.

As an illustration, consider the example in Figure \ref{fig:examples}.
In this case, we can separate the entire video into 11 scenes
as shown in Figure \ref{fig:examples}.
Accordingly, in the extended FOL structure, we have $S = \setof{s_1, \ldots, s_{11}}$.
Here the first scene, $s_1$, consists of the following:
the predicate \SR{run} holds of the entity $\mathsf{d_1}$,
the predicate \SR{sit} holds of the pair $(\mathsf{d_2}, x_1)$ where $x_1$ is an entity
which is a table. In terms of the interpretation function $I$ relativized to a scene, we have
$I_{s_1}(\SR{run}) = \setof{\mathsf{d_1}},
I_{s_1}(\SR{sit}) = \setof{(\mathsf{d_2},x_1)}$
and 
$I_{s_1}(\SR{table}) = \setof{x_1}$.
Similarly, we can extend the interpretation function $I$ to the other scenes.

While the triplet format is suitable for the annotation of various action labels,
the semantic representation in the form of
FOL structures with scenes can be directly used in
model checking and theorem proving for
the VTE system developed in \citet{suzuki2019}.
Our annotation format is flexible enough to be adapted in such applications.

\section{Conclusion}
We introduce a video-and-language dataset with human actions for multimodal inference.
We annotate human actions in videos in the free format and collect 1,942 action categories for 200 videos.
Our dataset contains various action labels for videos, including
those predicates characteristic of videos such as predicates for utterance,
predicates for intention and attitude,
and aspectual predicates.
In future work, we analyze recent action recognition models using Action Genome~\cite{action-genome} with our dataset.
We will also work on building a multimodal logical inference system between texts and videos.

\section*{Acknowledgment}

This work was partially supported by JST CREST Grant Number JPMJCR20D2, Japan.
\todo{Thanks to the anonymous reviewers for helpful comments.
We would also like to thank Mai Yokozeki and Natsuki Murakami for their contributions.}

\bibliographystyle{acl_natbib}
\bibliography{MyLibrary}

\end{document}